# Efficient Parametric Projection Pursuit Density Estimation

**Max Welling**
Dept. of Computer Science
University of Toronto
6 King's College Road
Toronto, M5S 3G4

**Richard S. Zemel**
Dept. of Computer Science
University of Toronto
6 King's College Road
Toronto, M5S 3G4

**Geoffrey E. Hinton**
Dept. of Computer Science
University of Toronto
6 King's College Road
Toronto, M5S 3G4

## Abstract

Product models of low dimensional experts are a powerful way to avoid the curse of dimensionality. We present the "under-complete product of experts" (UPoE), where each expert models a one dimensional projection of the data. The UPoE may be interpreted as a parametric probabilistic model for projection pursuit. Its ML learning rules are identical to the *approximate* learning rules proposed before for under-complete ICA. We also derive an efficient sequential learning algorithm and discuss its relationship to projection pursuit density estimation and feature induction algorithms for additive random field models.

## 1 INTRODUCTION

Projection pursuit density estimation (PPDE) is a sequential approach to train product models from data (Friedman et al., 1984). Each factor in the product is a "ridge function" which varies along one dimension and is constant along all other directions. The power of PPDE comes from the fact that it is able to largely avoid the curse of dimensionality by modelling one dimensional projections of the data and combining them in a multiplicative fashion. The disadvantage of PPDE is that it is computationally expensive.

In (Hinton, 1999) the "product of experts" (PoE) model and an efficient learning procedure (contrastive divergence) was introduced. More recent versions suited for the continuous domain model the data as over-complete products of 1-dimensional projections (Hinton and Teh, 2001; Teh et al., 2003). Although in some circumstances there are reasons to prefer over-complete representations, their model parameters are very hard to learn and approximate methods are needed. This paper introduces the under-complete PoE, which may be interpreted as a parametric probabilistic model for projection pursuit. Based on that relationship we derive an efficient and principled sequential learning algorithm in section 4.

Projection pursuit has seen many applications in data visualization, feature extraction, pattern recognition and data analysis. The parametric probabilistic model and efficient learning schemes proposed in this paper provide attractive alternatives to reach the same objectives.

Product models are certainly not new to the AI community. Different names have been invented for slight variations of the same theme. In the class of product models fall for instance additive random field models, log-linear models, exponential family models, maximum entropy models and square noiseless ICA models. The relation between a number of these models and the UPoE are described in section 6.

## 2 UNDER-COMPLETE PoE

Let $\mathbf{x} \in \mathbb{R}^D$ denote a random vector in a $D$ dimensional input space. Our model will consist of a number $J \leq D$ of "experts" modelling certain projections of the input space. We will denote a 1-dimensional expert with $\mathcal{T}(z_j|\alpha_j)$, where $z_j = \mathbf{w}_j^T \mathbf{x}$ is the projection of $\mathbf{x}$ onto the vector $\mathbf{w}_j$ and $\alpha_j$ represent additional parameters used to model the distribution of the expert. These experts are combined by taking their product. When $J < D$ this does not constitute a proper, normalizable probability distribution in $D$ dimensions. To repair this we fill the remaining dimensions (indicated by $\mathbf{y}$) with uncorrelated Gaussian noise. Assuming that we have preprocessed the data such that the mean is zero and the covariance matrix is equal to the identity (i.e. the data has been "sphered"), the model thus becomes,

$$p(\mathbf{y}, \mathbf{z}) = \prod_{i=1}^{D-J} \mathcal{N}(y_i|0, 1) \prod_{j=1}^{J} \mathcal{T}(z_j|\alpha_j) \qquad (1)$$

where $y_i = \mathbf{v}_i^T \mathbf{x}$ is a projection of $\mathbf{x}$ onto the vector $\mathbf{v}_i$. The vectors $\{\mathbf{v}_i\}$ form an orthonormal basis in the orthogonal complement of the space spanned by the vectors $\{\mathbf{w}_j\}$ which themselves need not be orthogonal nor normalized. If we collect the vectors $\{\mathbf{v}_i\}$ as rows in a matrix $\mathbf{V}$ and similarly for $\{\mathbf{w}_j\}$ in a matrix $\mathbf{W}$ we have the following relation,

$$\mathbf{V}^T\mathbf{V} = \mathbf{I} - \mathbf{W}^T(\mathbf{W}\mathbf{W}^T)^{-1}\mathbf{W} \quad (2)$$

where $\mathcal{P} = \mathbf{W}^T(\mathbf{W}\mathbf{W}^T)^{-1}\mathbf{W}$ is the projection matrix onto the space spanned by the vectors $\{\mathbf{w}_j\}$ and $\mathcal{P}^\perp = \mathbf{I} - \mathcal{P}$ is the projection matix onto the orthogonal complement of that space, i.e. the space spanned by the basis vectors $\{\mathbf{v}_i\}$.

Since the data are provided in $\mathbf{x}$ space we would like to express the probability density in those variables. Transforming variables in densities involves a volume factor (Jacobian) as follows,

$$p(\mathbf{x}) = p(\mathbf{y}, \mathbf{z}) \left| \frac{\partial(\mathbf{y}, \mathbf{z})}{\partial \mathbf{x}} \right| \quad (3)$$

where $|\cdot|$ is the absolute value of the determinant. Using the fact that $|\mathbf{A}| = \sqrt{|\mathbf{A}\mathbf{A}^T|}$, with $\mathbf{A}^T = [\mathbf{V}^T|\mathbf{W}^T]$, $\mathbf{V}\mathbf{W}^T = \mathbf{W}\mathbf{V}^T = 0$ and[1] $\mathbf{V}\mathbf{V}^T = \mathbf{I}$ we arrive at,

$$p(\mathbf{x}) = \prod_{i=1}^{D-J} \mathcal{N}(\mathbf{v}_i^T\mathbf{x}|0,1) \prod_{j=1}^{J} \mathcal{T}(\mathbf{w}_j^T\mathbf{x}|\alpha_j) \sqrt{|\mathbf{W}\mathbf{W}^T|} \quad (4)$$

We note that the model has no hidden variables and has a simple normalization constant given by $Z = 1/\sqrt{|\mathbf{W}\mathbf{W}^T|}$.

## 3 LEARNING

To learn the parameters of the UPoE model we chose the log-likelihood as our objective,

$$L = \frac{1}{2}\log|\mathbf{W}\mathbf{W}^T| + \quad (5)$$
$$\left\langle \sum_{i=1}^{D-J} \log \mathcal{N}(\mathbf{v}_i^T\mathbf{x}_n|0,1) + \sum_{j=1}^{J} \log \mathcal{T}(\mathbf{w}_j^T\mathbf{x}_n|\alpha_j) \right\rangle_{\tilde{p}}$$

where $\langle \cdot \rangle_{\tilde{p}}$ is an average with respect to the empirical data distribution $\tilde{p}$. To perform gradient ascent on this cost function, we need its derivatives with respect to the matrix $\mathbf{W}$ and the expert parameters $\{\alpha_j\}$,

$$\frac{\partial L}{\partial \mathbf{W}} = \mathbf{W}^{\#T} - \left\langle E'(\mathbf{z}) \mathbf{x}^T \right\rangle_{\tilde{p}} \quad (6)$$
$$\frac{\partial L}{\partial \alpha_j} = -\frac{\partial \log Z_j(\alpha_j)}{\partial \alpha_j} - \left\langle \frac{\partial E(z_j;\alpha_j)}{\partial \alpha_j} \right\rangle_{\tilde{p}} \quad (7)$$

where we have defined the pseudo-inverse by $\mathbf{W}^\# = \mathbf{W}^T(\mathbf{W}\mathbf{W}^T)^{-1}$. The "energy" $E(z_j)$ is defined through[2]

$$\mathcal{T}(z_j|\alpha_j) = \frac{1}{Z_j(\alpha_j)} e^{-E(z_j;\alpha_j)} \quad (8)$$

and $E'(\cdot)$ denotes its derivative. In the derivation of the learning rules we have used the fact that the quadratic noise term averaged over the (sphered) data is independent of the matrix $\mathbf{W}$,

$$\left\langle \mathbf{x}^T P^\perp \mathbf{x} \right\rangle_{\tilde{p}} = \mathrm{tr}\left(P^\perp \left\langle \mathbf{x}\mathbf{x}^T \right\rangle_{\tilde{p}}\right) = \mathrm{tr}\left(P^\perp\right) = D - J$$

with $\mathrm{tr}$ denoting the trace. We also used

$$\frac{1}{2}\frac{\partial \log|\mathbf{W}\mathbf{W}^T|}{\partial \mathbf{W}} = \mathbf{W}^{\#T} \quad (10)$$

Learning now proceeds through the updates $\mathbf{W} \to \mathbf{W} + \eta \partial L/\partial \mathbf{W}$ and $\alpha_j \to \alpha_j + \gamma \partial L/\partial \alpha_j$ for appropriate step-sizes $\eta$ and $\gamma$.

The learning rules for the matrix $\mathbf{W}$ in Eqn.6 have been reported before (Stone and Porrill, 1998; Lu and Rajapakse, 2000; Ridder et al., 2002), but as *approximate* learning rules for the under-complete ICA model (section 6.2). Here we have derived them as *exact* learning rules for the under-complete PoE model.

Assume for a moment that we are training the vector $\mathbf{w}_j$, while all other vectors $\{\mathbf{w}_1, ..., \mathbf{w}_{j-1}\}$ are kept fixed. It is important to observe that the first term in Eqn.6 depends on *all* vectors through the pseudo-inverse, while the second term only depends on the vector $\mathbf{w}_j$ that we are currently training. This implies that the first term represents what the model already knows about the data and therefore causes the vectors to diversify. Another way of seeing this is by rewriting the first term as,

$$\mathbf{W}^{\#T} = \left\langle E'(\mathbf{z}) \mathbf{x}^T \right\rangle_p \quad (11)$$

where $\langle \cdot \rangle_p$ means an average with respect to the model distribution $p$ (Eqn.4). One could choose to compute the above average through sampling from the model $p$ (see appendix A). The role of these samples is to shield off or "nullify" the data that are well represented by the model, causing learning to focus on poorly represented data. In fact, learning only stops when the average of $E'(\mathbf{z})\mathbf{x}^T$ over the empirical data distribution and the model distribution (represented by samples) match.

---

[1] Note that $\mathbf{V}^T\mathbf{V}$ is a projection operator which has $J - D$ eigenvalues 1, and the rest zeros. Assume that $\mathbf{u}$ is an eigenvector of $\mathbf{V}^T\mathbf{V}$ with eigenvalue 1, then $\mathbf{V}\mathbf{u}$ is an eigenvector of $\mathbf{V}\mathbf{V}^T$ with eigenvalue 1. Hence, all eigenvalues of $\mathbf{V}\mathbf{V}^T$ are 1 which implies that it must be equal to $\mathbf{I}$.

[2] Note that the normalizing constant for each individual expert depends on $\alpha_j$ but not on $\mathbf{w}_j$.

## 4 SEQUENTIAL LEARNING

Learning the vectors $\{\mathbf{w}_j\}$ can occur either in parallel or sequentially. However, since the calculation of the pseudo-inverse $\mathbf{W}^\#$ is the computational bottleneck, the latter doesn't seem a very attractive option[3]. What is needed is a way to avoid the recomputation of $\mathbf{W}^\#$ for every step of gradient ascent learning. In the following we will propose a more efficient sequential learning algorithm based on projection pursuit density estimation (PPDE) (Friedman et al., 1984; Huber, 1985).

In PPDE the learning procedure is split into two parts: in one phase we search for a direction $\hat{\mathbf{w}}$ in which the projected data look non-normally distributed, i.e. we look for "interesting" directions in the data. This is implemented by defining a projection index and minimizing it. In the other phase, we fit a model to the marginal distribution in that direction and use it to replace the current estimate of that marginal distribution. In fact, one can show that for a given direction $\hat{\mathbf{w}}_j$, the optimal multiplicative update for the model at round $j-1$ is given by,

$$p_j(\mathbf{x}) = p_{j-1}(\mathbf{x}) \frac{p_{\text{data}}^{\hat{\mathbf{w}}_j}(\hat{\mathbf{w}}_j^T \mathbf{x})}{p_{j-1}^{\hat{\mathbf{w}}_j}(\hat{\mathbf{w}}_j^T \mathbf{x})} \quad (12)$$

where $p_{\text{data}}^{\hat{\mathbf{w}}_j}$ is the marginal data distribution in the direction $\hat{\mathbf{w}}_j$ and $p_{j-1}^{\hat{\mathbf{w}}_j}$ is the marginal model distribution at round $j-1$ in direction $\hat{\mathbf{w}}_j$. Note that the new model distribution $p_j(\mathbf{x})$ is still normalized after this update. This procedure is initiated with a "prior" noise model $p_0(\mathbf{x})$ which is typically a multivariate standard normal distribution.

It not difficult to compute the change in the Kullback-Leibler divergence between the data and the model distribution due to this update,

$$\begin{aligned} \mathcal{Q} &= \mathcal{D}\left(p_{\text{data}}\|p_j\right) - \mathcal{D}\left(p_{\text{data}}\|p_{j-1}\right) \\ &= -\mathcal{D}\left(p_{\text{data}}^{\hat{\mathbf{w}}_j}\|p_{j-1}^{\hat{\mathbf{w}}_j}\right) \end{aligned} \quad (13)$$

PPDE *minimizes* this projection index $\mathcal{Q}$ over directions $\hat{\mathbf{w}}_j$. The algorithm thus searches for directions for which the improvement of the model is largest. These are the directions where the model is most different from the data distribution. Although theoretically appealing, the computational load of this algorithm is large. This is due to the fact that the marginal distributions are typically modelled by splines or histograms, which makes the computation of the KL divergence $\mathcal{D}\left(p_{\text{data}}^{\hat{\mathbf{w}}_j}\|p_{j-1}^{\hat{\mathbf{w}}_j}\right)$ and its derivatives (needed for gradient descent) cumbersome.

---

[3]One can define a *natural gradient*. However, unlike the complete (square) case, this natural gradient still depends on the pseudo-inverse, therefore not resulting in improved efficiency per iteration. Because we work with sphered data the covariant form of the updates will also not lead to faster convergence.

We will now describe a procedure, based on the above ideas, that trains the UPoE model sequentially. Due to the parametric form of the experts, this learning algorithm will turn out to be very efficient albeit less flexible than the plain vanilla PPDE procedure. We first observe that the addition of an expert to the UPoE model in a direction orthogonal to the previous experts can be written as follows,

$$p_j(\mathbf{x}) = p_{j-1}(\mathbf{x}) \frac{\mathcal{T}_j(\hat{\mathbf{w}}_j^T \mathbf{x}; \alpha_j)}{\mathcal{N}(\hat{\mathbf{w}}_j^T \mathbf{x})} \quad (14)$$

This is precisely in the form of Eqn.12 if the marginal data distribution in the direction $\hat{\mathbf{w}}_j$ is perfectly described by the expert $\mathcal{T}_j$. Note that the fact that the model $p_{j-1}(\mathbf{x})$ is indeed normal in any direction orthogonal to the vectors $\{\mathbf{w}_1, ..., \mathbf{w}_{j-1}\}$ guarantees that the new model $p_j$ is again normalized. To compute the projection index we determine the change in KL divergence between the data distribution and the model distribution, assuming update Eqn.14,

$$\begin{aligned} \mathcal{Q} &= \mathcal{D}\left(p_{\text{data}}\|p_j\right) - \mathcal{D}\left(p_{\text{data}}\|p_{j-1}\right) \quad (15) \\ &= \mathcal{D}\left(p_{\text{data}}^{\hat{\mathbf{w}}_j}\|\mathcal{T}_j\right) - \mathcal{D}\left(p_{\text{data}}^{\hat{\mathbf{w}}_j}\|\mathcal{N}\right) \quad (16) \end{aligned}$$

In contrast to the PPDE objective in Eqn.13 this projection index has *two* terms. The first term searches for directions in which the data can be well described by the expert while the second term prefers directions that are "interesting", i.e. non-normally distributed. Compared to Eqn.13, the first term is new and appears because we didn't assume that the data distribution in the direction $\hat{\mathbf{w}}_j$ can be modelled with infinite precision. However, the most important difference between the two projection indices is that the latter is computationally more efficient than the one used for PPDE. This can be seen by inserting the empirical data distribution $\tilde{p}$ for $p_{\text{data}}$ and rewriting Eqn.16 as,

$$\begin{aligned} \mathcal{Q} = & \left\langle E_j(\hat{\mathbf{w}}_j^T \mathbf{x}; \alpha_j) - \frac{1}{2}(\hat{\mathbf{w}}_j^T \mathbf{x})^2 \right\rangle_{\tilde{p}} \\ & + \log Z_j(\alpha_j) - \frac{1}{2}\log(2\pi) \end{aligned} \quad (17)$$

which is trivially evaluated, as are its derivatives w.r.t. $\hat{\mathbf{w}}_j$ given by

$$\frac{\partial \mathcal{Q}}{\partial \hat{\mathbf{w}}_j} = \left\langle \left(E_j'(\hat{\mathbf{w}}_j^T \mathbf{x}; \alpha_j) - \hat{\mathbf{w}}_j^T \mathbf{x}\right) \mathbf{x}^T \right\rangle_{\tilde{p}} \quad (18)$$

There are two reasons for the simplification. Firstly, the entropy of the marginal data distribution $p_{\text{data}}^{\tilde{\mathbf{w}}}$ drops out of the projection index. Secondly, we force the model $p_{j-1}$ to be normally distributed in the new direction by choosing it orthogonal to the previous directions, resulting in the simple update Eqn.14. To maintain that condition we need to re-orthogonalize $\hat{\mathbf{w}}_j$ with respect to $\{\hat{\mathbf{w}}_1, ..., \hat{\mathbf{w}}_{j-1}\}$ after every gradient update,

$$\mathbf{w}_j \to \mathbf{w}_j - \mathbf{W}_{j-1}^T \mathbf{W}_{j-1} \mathbf{w}_j \qquad \mathbf{w}_j \to \mathbf{w}_j / \|\mathbf{w}_j\| \qquad (19)$$

where $\mathbf{W}_{j-1}$ is the matrix with the vectors $\{\mathbf{w}_1, ..., \mathbf{w}_{j-1}\}$ as its rows.

The expert parameters $\alpha_j$ can be learned *simultaneously* with the directions $\hat{\mathbf{w}}_j$, by minimizing $\mathcal{Q}$. The gradients are given by,

$$\frac{\partial \mathcal{Q}}{\partial \alpha_j} = \left\langle \frac{\partial E_j(\hat{\mathbf{w}}_j^T \mathbf{x}; \alpha_j)}{\partial \alpha_j} \right\rangle_{\tilde{p}} + \frac{\partial \log Z_j(\alpha_j)}{\partial \alpha_j} \qquad (20)$$

which is precisely equivalent to Eqn.7. This flexibility to adapt the expert at the same time as we learn the direction $\hat{\mathbf{w}}_j$ may become important if different directions in the data have qualitatively different marginals, e.g. some directions could be highly kurtotic while others could be bimodal.

Since $\mathcal{Q}$ represents the change in the negative log-likelihood (if we replace the data distribution with the empirical distribution) we should stop learning when all remaining directions satisfy $\mathcal{Q} \geq 0$.

Finally, we summarize the resulting algorithm below:

---
**Sequential Learning Algorithm for UPoEs**

---
*Repeat for $j=1$ to $J$ or until $\mathcal{Q} \geq 0$:*
1. Initialize $\hat{\mathbf{w}}_j$ to a random unit length D-vector, orthogonal to the previous directions $\{\hat{\mathbf{w}}_1, ..., \hat{\mathbf{w}}_{j-1}\}$.
2. *Repeat until convergence:*
    2i. Take a gradient step to decrease projection index $\mathcal{Q}$ over directions $\hat{\mathbf{w}}_j$ (Eqn.18) and parameters $\alpha_j$ (Eqn.20).
    2ii. Apply Gram-Schmidt orthonormalization to $\hat{\mathbf{w}}_j$ (Eqn.19).
3. Update model (Eqn.14).

---

## 5 EXPERIMENTS

To compare the parallel learning algorithm of section 3 with the sequential algorithm described in section 4 we trained models with varying number of projections. The data-set[4] consisted of 1965 face images of

[4]Obtained from *www.cs.toronto.edu/~roweis/data*

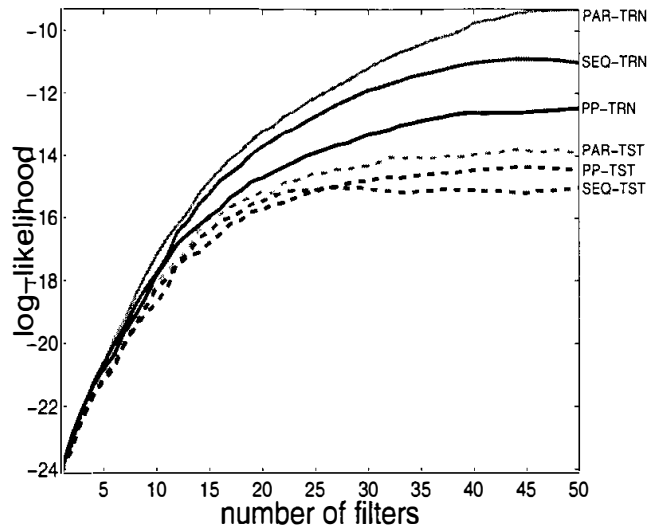

Figure 1: Models trained on the "Frey-faces" data set with varying numbers of projections. Solid curves represent log-likelihood on training data while dashed lines represent log-likelihood on test data. "PAR" indicates that the parameters were trained in parallel using Eqn.6. "SEQ" indicates that the model was trained using Eqn.6 sequentially (see main text for details) and "PP" indicates that the model was trained using the algorithm described in section 4.

128 pixels each. This data-set was centered, sphered and reduced to 50 dimensions using PCA, keeping only the high variance directions. The data cases were split into 1000 randomly chosen training cases and 965 test cases. We used Student-t experts (see appendix B) to describe the marginal distributions which can gracefully interpolate between a normal distribution and a super-Gaussian distribution (highly peaked distribution with heavy tails).

In figure 1 we show the log-likelihood for three different training procedures. The (green) curve indicated with "PAR-TRN" shows the results for the parallel update (Eqns.6,7). Each time a vector $\mathbf{w}_j$ is added to the model, the other vectors are initialized at the ones previously learned, but are allowed to change during learning. The dashed (green) line indicated by "PAR-TST" shows the result on the test data. The (blue) curves indicated by "PP-TRN" and "PP-TST" are the training and testing results for the sequential learning algorithm described in section 4. The (red) curves indicated by "SEQ-TRN" and "SEQ-TST" show the results for a procedure very similar to the parallel algorithm, but without the ability to update the previously learned parameters. Although the parallel procedure outperforms the sequential methods on training data, there is no significant difference on the test data.

In another experiment we used a mixture of 2 Student-t distributions (see appendix B) with fixed settings of the inverse temperature at $\beta_1 = \beta_2 = 20$ and means at $\mu_1 = -1, \mu_2 = +1$ but with adaptable scale parameters $\{\theta_1, \theta_2\}$ and mixture coefficients $\{\pi_1, \pi_2\}$. The goal of this experiment was to verify that our projection pursuit algorithm could extract the interesting multi-modal projections from the " Leptograpsus Crabs" data set[5] . This data set contains 50 specimens of each sex of two color forms (Ripley, 1996). The Crabs data were first centered and sphered before presentation to the sequential learning algorithm. Figure 2 and 3 show the results. It was found that there were many local minima present and each time the results looked slightly different.

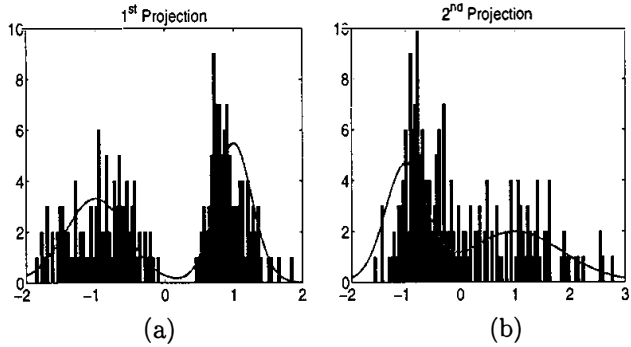

Figure 2: First projection (a) and second projection (b) of the Crabs data set found by the sequential learning algorithm. Overlayed are the fitted mixture of Student-t distributions with means fixed at $\mu_1 = -1$ and $\mu_2 = +1$ (solid red curve).

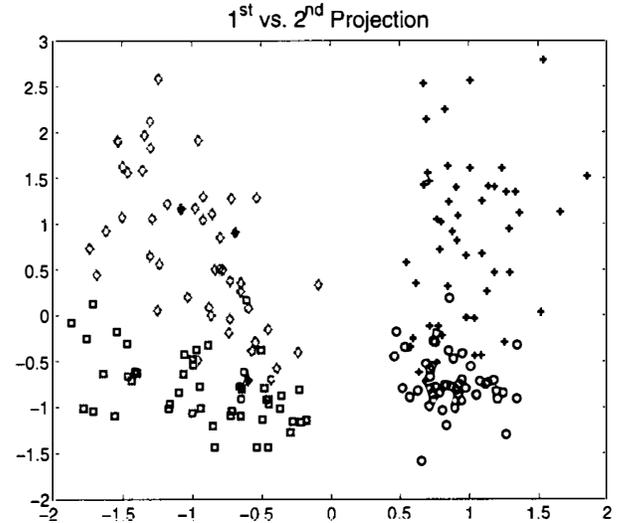

Figure 3: First versus second projection of the Crabs data set corresponding to the histograms shown in figure 2. The 4 different symbols ($\circ, \square, \lozenge, +$) correspond to the 4 different classes of crab species and sex. Note however that these labels were not used by the algorithm to determine the projections.

Finally we collected 100,000 patches of natural images[6] of size $30 \times 30$ pixels. This data set was centered, sphered and reduced to 400 dimensions using PCA. Subsequently, 100 directions $\hat{\mathbf{w}}$ were learned using the sequential algorithm. Training was done in batches of 100 cases, involved an adaptive step-size and took a few hours on a 1-Gz PC. In figure 4 we show 25 randomly chosen "filters" (rows of $WA_{\text{pca}}$). The results are qualitatively similar to the Gabor-like receptive fields found using ICA in (Bell and Sejnowski, 1997).

## 6 RELATION TO OTHER MODELS

### 6.1 PROJECTION PURSUIT

In section 4 we have described the forward or synthetic PPDE procedure. The resulting model is in fact very similar to a UPoE model, with the subtle difference that the background model in PPDE is a full dimensional standard normal distribution while the background model for a UPoE is normal only in the orthogonal complement of the space spanned by the projections. The PPDE model is very flexible because the marginals are fit with histograms or splines and the projections are not necessarily orthogonal. However, relative to the parametric UPoE, the PPDE procedure is computationally inefficient.

There is also a backward or analytic approach to PPDE (Friedman, 1987). The idea is that one starts with the data distribution and sequentially strips away its non-Gaussian structure. A practical method to inform the algorithm about the structure that has already been detected in previous rounds is to "Gaussianize" those directions. This amounts to transforming the old data set into a new one where previously detected directions are now normally distributed. This technique only works when the directions are mutually orthogonal. In (Zhu, 2002) it is argued that the resulting density model is in fact very cumbersome, due to the need to "unwrap" these transformations.

### 6.2 UNDER-COMPLETE ICA

Probabilistic models for independent components analysis take the form of causal (directed) models where a number of sources $\{s_i\}$, distributed according to the prior distributions $p_i(s_i)$, are linearly combined to produce the observed random variables $\mathbf{x}$. The model is given by,

$$p(\mathbf{x}) = p(\mathbf{x}|\mathbf{s}) \prod_{i=1}^{J} p_i(s_i) \qquad (21)$$

---
[5]Obtained from www.stats.ox.ac.uk/pub/PRNN/
[6]Obtained from www.cis.hut.fi/projects/ica/data/images

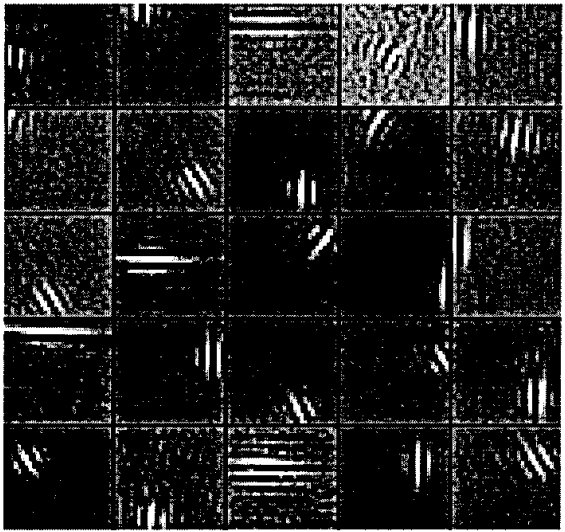

Figure 4: 25 randomly chosen projections out of a 100 learned projections (using the sequential algorithm) on the natural image data set.

where $p(\mathbf{x}|\mathbf{s})$ is the likelihood term, which models the noise of the observed variables. A noise model is necessary in the under-complete setting to make the probability distribution proper, i.e. normalizable over input space.

This model is different from the under-complete PoE model since it is defined in terms of stochastic hidden source variables $\{s_i\}$ which are difficult to integrate out. Along the lines of the discussion presented in (Teh et al., 2003), the UPoE model is a "filter model", while the under-complete ICA model is a causal generative model. However, the two models are closely related as is evidenced by the fact that the learning rule using the derivative Eqn.6 has been proposed as an approximation to the intractable gradient ascent learning rule for the under-complete ICA model (Stone and Porrill, 1998; Lu and Rajapakse, 2000; Ridder et al., 2002). In the complete noiseless case, i.e. when $J = D$ and $p(\mathbf{x}|\mathbf{s}) = \delta(\mathbf{x} - \mathbf{W}^{-1}\mathbf{s})$, the PoE and ICA models become in fact equivalent. This implies that the presented sequential learning algorithm, when completed until $J = D$, can be interpreted as a sequential learning algorithm for square noiseless ICA.

ICA does not hinge on the existence of a probabilistic model. The initial formulations used objectives such as mutual information and negentropy of the linearly transformed input variables, or mutual information between input variables and non-linearly transformed input variables. All these formulations, including the probabilistic approach, turn out to be related under certain conditions (Hyvarinen and Oja, 2000). The formulation that is closest in spirit to the one presented in this paper is the one underlying the fastICA algorithm (Hyvarinen, 1999). There, new directions are added sequentially by minimizing the negentropy as a projection index. The main difference with the sequential procedure described here is that our (different) projection index is based on the maximal decrease in log-likelihood of a probabilistic model.

### 6.3 ADDITIVE RANDOM FIELD MODELS

In the discrete domain product models are known under various names; additive random field models (Pietra et al., 1997), log-linear models (Darroch and Ratcliff, 1972) and maximum entropy models (Zhu et al., 1997). These models have an elegant dual interpretation as the distribution that minimizes the KL divergence $\mathcal{D}(p||p^0)$ with a "prior model" $p^0$, subject to a number of constraints $\langle \Phi_i(\mathbf{x}) \rangle_{p_{\text{model}}} = \langle \Phi_i(\mathbf{x}) \rangle_{p_{\text{data}}}$. The resulting model has the form,

$$p(\mathbf{x}) \propto p^0(\mathbf{x}) e^{\sum_{i=1}^{J} \lambda_i \Phi_i(\mathbf{x})} \qquad (22)$$

The features $\Phi_i$ are selected from a large "library" while the Lagrange multipliers $\lambda_i$ (or weights), which multiply the features are learned. This procedure is typically sequential, minimizing a similar objective as the projection index proposed in this paper, $\mathcal{D}(p_{\text{data}}||p_j) - \mathcal{D}(p_{\text{data}}||p_{j-1})$. Identifying features with $\Phi(y_i, \mathbf{x}) = \delta(y_i - \hat{\mathbf{w}}_i^T \mathbf{x})$ we can show that the optimal choice for the weights is given by $\lambda(y_i) = \log \mathcal{T}(y_i; \alpha_i) - \log \mathcal{N}(y_i)$, which precisely corresponds to the UPoE model.

The discrete search over features is therefore replaced in our case with a continuous optimization over directions $\hat{\mathbf{w}}$ while the estimation of the weights $\lambda$ could be identified with the optimization of expert parameters $\alpha$. Thus, in many respects the UPoE model and its sequential learning algorithm are the analogue in the continuous domain of the additive random field model and its feature pursuit algorithm.

## 7  CONCLUSION

The UPoE model and its learning algorithms provide a link between under-complete ICA, projection pursuit and additive random field models. The parallel learning rules have been proposed in the literature as approximate learning rules for under-complete ICA. This paper provides insight into what those learning rules really accomplish. The sequential learning rules can be interpreted as a a parametric variant of PPDE, but are also similar in spirit to the fastICA fitting method. In fact, when the number of experts is equal to the dimensionality of the input space it constitutes an ICA learning algorithm. Finally, the UPoE and its sequential learning rules may be interpreted as the continu-ous analogue of additive random field models and their feature induction techniques.

Important features of the UPoE are its simplicity and its efficient learning rules. The most important disadvantage is that the story breaks down if the number of experts exceeds the number of input dimensions. In some situations there are reasons to prefer over-complete PoE models, but it turns out that they are much harder to learn (Teh et al., 2003).

Another limitation is its restriction to the continuous domain. Many applications, such as document retrieval and language processing, require models to work in the (positive) discrete data domains. Many existing models, such as the aspect model, suffer from intractable inference which is needed for learning. Extending PoE models into this domain is a topic of future research.

In (Welling et al., 2002) a product model was described that has the ability to topographically order its projections. This idea readily extends to the UPoE model. Unfortunately, learning has to proceed using approximate methods such as contrastive divergence. Extending these ideas to the discrete domain may have interesting applications in latent semantic analysis where the topics can be ordered topographically according to their interdependencies.

**Acknowledgements**

We would like to thank Yee Whye Teh, Simon Osindero and Sam Roweis for discussions and suggestions.

## A  SAMPLING

Sampling from the UPoE is relatively straightforward. We sample $z_j \sim \mathcal{T}(z_j|\alpha_j)$ and $y_i \sim \mathcal{N}(y_i|0,1)$ and combine them into a sample in **x**-space using,

$$\mathbf{x} = \mathcal{P}\mathbf{x} + \mathcal{P}^\perp \mathbf{x} = \mathbf{W}^\# \mathbf{z} + \mathbf{V}^T \mathbf{y} \qquad (23)$$

To explicitly compute an orthonormal basis $\mathbf{V}^T$ we can compute the following SVD decomposition $\mathbf{ABC}^T = SVD([\mathbf{W}^T|\mathbf{0}])$. The last $J - D$ columns of $\mathbf{A}$ then form and orthonormal basis in the complement subspace, $\mathbf{A} = [\mathbf{A}_{DJ}|\mathbf{V}^T]$. Moreover, the pseudo-inverse of $\mathbf{W}$ can also be computed as $\mathbf{W}^\# = A_{DJ}\mathbf{B}_{JJ}^{-1}\mathbf{C}_{JJ}^T$. Alternatively, we can sample $\mathbf{x}' \sim \mathcal{N}(\mathbf{x}'|0,\mathbf{I})$ and subsequently project the samples on the orthogonal subspace: $\mathbf{V}^T \mathbf{y} = \mathcal{P}^\perp \mathbf{x}'$.

## B  STUDENT-T EXPERTS

The probability distribution of a (generalized) Student-t distribution is given by,

$$\mathcal{T}(z) = \frac{\Gamma(\beta)\,\theta}{\Gamma(\beta - \frac{1}{2})\sqrt{2\pi}} \left(1 + \frac{1}{2}(\theta(z-\mu))^2\right)^{-\beta} \qquad (24)$$

where $\mu$ is its mean, $\theta > 0$ is an inverse scale parameter and $\beta > \frac{1}{2}$ an inverse "temperature" which controls the sharpness of the distribution. We can easily sample from it for arbitrary $\beta$ and $\theta$: first compute $a = \beta - \frac{1}{2}$ and $b = \theta^2$. Next, sample precision parameters from a gamma distribution, $y \sim c\, y^{a-1} e^{-y/b}$. Finally, sample $z$ from a normal distribution with that precision $z \sim c\, e^{-\frac{1}{2}y\, z^2}$. The derivatives of the log-likelihood for the parameters $\mu, \theta, \beta$ are given by,

$$\frac{\partial L}{\partial \theta} = \frac{1}{\theta} - \left\langle \frac{\beta \theta z^2}{1 + \frac{1}{2}(\theta z)^2} \right\rangle_{\tilde{p}} \qquad (25)$$

$$\frac{\partial L}{\partial \beta} = \Psi(\beta) - \Psi\left(\beta - \frac{1}{2}\right) - \left\langle \log\left(1 + \frac{1}{2}(\theta z)^2\right) \right\rangle_{\tilde{p}} \qquad (26)$$

$$\frac{\partial L}{\partial \mu} = \left\langle \frac{-\beta \theta^2 (z-\mu)}{1 + \frac{1}{2}(\theta(z-\mu))^2} \right\rangle_{\tilde{p}} \qquad (27)$$

where $\Psi(x) = \partial/\partial x \ln \Gamma(x)$ is the "digamma" function. It is not hard to compute the variance and kurtosis of a central Student-T distribution,

$$\langle z^2 \rangle_{\mathcal{T}} = \frac{1}{\theta^2 (\beta - \frac{3}{2})} \qquad \beta > \frac{3}{2} \qquad (28)$$

$$\frac{\langle z^4 \rangle_{\mathcal{T}}}{\langle z^2 \rangle_{\mathcal{T}}} - 3 = \frac{3}{(\beta - \frac{5}{2})} \qquad \beta > \frac{5}{2} \qquad (29)$$

Thus, small values for $\beta$ represent peaked distributions with large kurtosis. The derivative of the projection index with respect to the direction $\hat{\mathbf{w}}_j$ is given by,

$$\frac{\partial Q}{\partial \hat{\mathbf{w}}_j} = \left\langle \left( \frac{\beta_j \theta_j^2 \hat{\mathbf{w}}_j^T \mathbf{x}}{1 + \frac{1}{2}(\theta_j \hat{\mathbf{w}}_j^T \mathbf{x})^2} - \hat{\mathbf{w}}_j^T \mathbf{x} \right) \mathbf{x}^T \right\rangle_{\tilde{p}} \qquad (30)$$

A more general family of experts is given by *mixtures* of Student-t distributions.

$$\mathcal{T}(\mathbf{x}) = \sum_a \pi_a \mathcal{T}_a(\mathbf{x}; \alpha_a) \qquad (31)$$

where $\pi_a$ are the mixture coefficients. Learning mixture models is straightforward using the EM algo-rithm,

$$r_{an} = \frac{\pi_a \mathcal{T}_a(z_n; \alpha_a)}{\sum_a \pi_a \mathcal{T}_a(z_n; \alpha_a)} \quad (32)$$

$$\pi_a^{\text{new}} = \frac{1}{N} \sum_n r_{an} \quad (33)$$

$$\alpha_a^{\text{new}} = \alpha_a + \frac{\varepsilon}{N} \sum_n r_{an} \frac{\partial \log \mathcal{T}_a(z_n; \alpha_a)}{\partial \alpha_a} \quad (34)$$

$$\hat{\mathbf{w}}^{\text{new}} = \hat{\mathbf{w}} + \frac{\varepsilon}{N} \sum_a \sum_n r_{an} \frac{\partial \log \mathcal{T}_a(z_n; \alpha_a)}{\partial \hat{\mathbf{w}}_a} \quad (35)$$

where the projections also need to be orthonormalized at every step. For $\mu$ and $\theta$ there are faster IRLS updates available (Titterington et al., 1985),

$$w_{an} = \frac{r_{an}}{\left(1 + \frac{1}{2}(\theta_a(z_n - \mu_a))^2\right)} \quad (36)$$

$$\mu_a^{\text{new}} = \frac{\sum_n w_{an} z_n}{\sum_n w_{an}} \quad (37)$$

$$(\theta^2)_a^{\text{new}} = \frac{N \pi_a^{\text{new}}}{\sum_n \alpha_a w_{an}(z_n - \mu_a)^2} \quad (38)$$

The new weights $w_{an}$ downweight outliers which makes this a robust alternative to the mixture of Gaussians model (Titterington et al., 1985).